\pdfoutput=1

\documentclass[11pt,a4paper]{article}
\usepackage[hyperref]{acl2021}

\usepackage{times}
\usepackage{latexsym}

\usepackage[T1]{fontenc}

\usepackage[utf8]{inputenc}

\usepackage{microtype}

\aclfinalcopy 

\usepackage{inconsolata}

\usepackage{soul}
\usepackage{booktabs}
\usepackage{lipsum}
\usepackage{multirow}
\usepackage{xurl}

\usepackage{tikz} \usetikzlibrary{calc, arrows.meta, intersections, patterns, positioning, shapes.misc, fadings, through,decorations.pathreplacing}
\usetikzlibrary{trees,positioning,shapes,shadows,arrows.meta}

\usepackage{dirtree}

\newcommand{\fespace}[1]{\framebox[1.9cm][l]{#1}}
\newcommand{\entity}[1]{\underline{#1}}

\def\checkmark{\tikz\fill[scale=0.4](0,.35) -- (.25,0) -- (1,.7) -- (.25,.15) -- cycle;} 
\def\rate{RaTE}

\title{\rate: a Reproducible automatic Taxonomy Evaluation by Filling the Gap}

\author{Tianjian Gao, Philippe Langlais \\
  RALI/DIRO, Université de Montréal \\  
  \texttt{tianjian.gao@umontreal.ca,felipe@iro.umontreal.ca}}

\begin{document}
\maketitle

\begin{abstract}
Taxonomies are an essential knowledge representation, 
yet most studies on automatic taxonomy construction (ATC) resort to manual evaluation to score proposed algorithms. 
We argue that automatic taxonomy evaluation (ATE) is just as important as taxonomy construction. 
We propose \rate{,}\footnote{Our code repository is available at \url{https://github.com/CestLucas/RaTE}.} an automatic label-free taxonomy scoring procedure, which relies on a large pre-trained language model. 
We apply our evaluation procedure to three state-of-the-art ATC algorithms with which we built seven taxonomies from the Yelp domain, and show that 1) 
\rate{} correlates well with human judgments and 2) artificially degrading a  taxonomy leads to decreasing \rate{} score.

\end{abstract}

\section{Introduction}

A domain taxonomy is a tree-like structure
that not only aids in knowledge organization but also serves an integral part of many knowledge-rich applications 
including web search, recommendation systems and  decision making processes. 
Taxonomies are also inevitably used as business and product catalogs and for managing online sales. 
Notable taxonomy products in this domain include Amazon Category Taxonomy,\footnote{\url{https://www.data4amazon.com/amazon-product-taxonomy-development-mapping-services.html}}
Google Product Taxonomy,\footnote{\url{https://support.google.com/merchants/answer/6324436?hl=en}} Yelp Business Category\footnote{\url{https://blog.yelp.com/businesses/yelp_category_list/}} and Google Content Categories.\footnote{\url{https://cloud.google.com/natural-language/docs/categories?hl=fr}}

Recent years have witnessed interest in new automatic taxonomy construction (ATC) systems, 
but there are no systematic methods for objectively evaluating their figure of merit. 
For instance, TaxoGen \cite{zhang2018taxogen} --- see Section~\ref{sec:system_description} --- 
was evaluated by asking at least three human evaluators 
if a taxonomy concept pair contains a hypernymy relationship, which can lead to bias and low reproducibility.
It is not only difficult to compare or rank different algorithms, 
but changing the hyper-parameters or settings of a parameterized ATC system can also result in drastically different outputs, 
and make optimization unfeasible.

Because ontologies and taxonomies in particular are typically created in contexts to address specific problems or achieve specific goals,
e.g. classification, their evaluation is evidently context-dependent, and many researchers actually believe that a task-independent 
automatic evaluation remains elusive  \cite{porzel2004task}.
Still, researchers have argued that objective evaluation metrics must be well available for 
significant progress in the development and deployment of taxonomies and ontologies \cite{Brewster2004}.


In this work, we propose \rate, a Reproducible procedure for 
Automatic Taxonomy Evaluation. \rate{} does not require external knowledge but instead 
depends on masked language modelling (MLM) to query a large language model for  subsumption relations.  We show that with some care, MLM  is a valuable proxy to human judgments. 

We apply \rate{}  to the Yelp corpus (a corpus of restaurant  reviews) ranking seven taxonomies we extracted using three state-of-the-art ATC systems. We observe it correlates well with our manual evaluation of those taxonomies, and also show that artificially degrading a taxonomy leads to a decrease of score proportional to the level of noise injected.

In the remainder, we discuss related work in Section~\ref{sec:related}.  In Section~\ref{sec:system_description}, we describe
the ATC systems we used for building up our taxonomies, 
and their evaluation procedures.  
We then present \rate{} in Section~\ref{sec:approach} including refinements that 
we found necessary for our approach to work.  We report in Section~\ref{sec:experiments} the experiments we conducted to demonstrate the relevance  of \rate, and conclude in Section~\ref{sec:discussion}.

\section{Related Works}
\label{sec:related}

Systematic methods of evaluating ontologies and taxonomies are lacking.  Because agreed upon quantitative metrics are lacking, 
research on taxonomy and ontology construction relies heavily
on qualitative descriptions and the various perspectives of ontology engineers, 
system users or domain experts, which renders the results subjective and unreproducible \cite{gomez1999evaluation,guarino1998some}.

\citet{brank2005survey} 
summarized four principle ontology evaluation methods, by 
(1) comparing the target ontology to a "gold standard" (ground-truth) ontology \cite{maedche2002measuring}; 
(2) using the target ontology in an application and evaluating the application results ("application based") \cite{porzel2004task}; 
(3) conducting coverage analysis comparing the target with a source of data (eg., a collection of documents) 
about a specific domain ("data driven") \cite{Brewster2004}; 
(4) manual reviews done by human experts that assess how well the target ontology meets a set of predefined criteria, standards, 
and requirements \cite{lozano2004ontometric}.

\paragraph{Gold Standard Evaluation}  focusses on comparing and measuring the similarity of the target taxonomy with an existing ground truth such as WordNet \cite{fellbaum98wordnet}, Wikidata and ResearchCyc \cite{journals/ai/PonzettoS11}. 
Semantic similarity metrics have been proposed, including Wu-Palmer \cite{wu1994verb}, Leacock-Chodorow 
\cite{leacock1998combining} and Lin \cite{lin1998information}. We include in this category specific measures such as  \emph{topic coherence}  \cite{newman2010automatic} which scores the quality of a word cluster which rely on similarity measures.  There are several issues with such a process: mapping concepts from the output system to the ground truth is not trivial and gold standards do not necessarily cover well the domains of interest.

\paragraph{Application-based Evaluation} is an attractive alternative
to gold-standard evaluation.  \citet{porzel2004task} for instance proposed several possible applications for evaluation including concept-pair relation classification.  \citet{brank2005survey} underlines however that it is in fact hard to correlate ontology quality with the application performance.

\paragraph{Data-driven Evaluation} intends to select the ontology $O$ with the best structural \emph{fit} 
to a target corpus $C$, which boils down into estimating $P(C|O)$ as in \cite{Brewster2004}. 
Practically however, it remains unclear how to approximate such conditional probability.

\section{Automatic Taxonomy Extractors}
\label{sec:system_description}

In this work, we replicated results of three state-of-the-art ATC systems that are publicly available
and that are producing quality results on selected datasets and domains. 
In this section, we describe those systems and discuss their corresponding evaluation methods.
  
\subsection{TaxoGen}
TaxoGen \cite{zhang2018taxogen} is an adaptive text embedding and clustering algorithm leveraging 
various phrase-mining and clustering techniques including AutoPhrase \cite{shang2018automated}, 
caseOLAP \cite{liem2018phrase} and spherical k-means clustering \cite{banerjee2005clustering}. 
TaxoGen iteratively refines selected keywords and chooses cluster representative terms based on two criteria:
\emph{popularity} which prefers term-frequency in a cluster and \textit{concentration} which assumes that 
representative terms should be more relevant to their belonging clusters than their sibling clusters. 

The system can be configured with several hyper-parameters including 
the depth of the taxonomy, the number of children per parent term and the "representativeness" threshold.
Experiments were conducted on DBLP and SP (Signal Processing) datasets and the system is quantitatively evaluated 
with relation accuracy and term coherency measures assessed by human evaluators (10 doctoral students).

\subsection{CoRel}
CoRel \cite{huang2020corel} takes advantages of novel relation transferring and concept learning techniques 
and uses hypernym-hyponym pairs provided in a seeded taxonomy to train a BERT \cite{devlin2018bert} relation classifier and expand the seeded taxonomy horizontally (width expansion) and vertically (depth expansion).
Topical clusters are generated using pre-computed BERT embeddings and a discriminative embedding space is learned, 
so that each concept is surrounded by its representative terms. 

The clustering algorithms used by CoRel are \emph{spectral co-clustering} \cite{kluger2003spectral} and
\emph{affinity propagation} \cite{frey2007clustering}, 
which automatically computes the optimal number of topic clusters.
Compared to TaxoGen, CoRel does not require depth and cluster number specifications but a 
small seeding taxonomy as an input for enabling a weakly-supervised relation classifier.  

CoRel is quantitatively evaluated with term coherency, relation F1 and sibling distinctiveness judged by 
5 computer science students on subsets of DBLP and Yelp datasets.
The system generates outputs in the form of large hierarchical topic word clusters.

\subsection{HiExpan}

HiExpan \cite{journals/corr/abs-1910-08194} is a hierarchical tree expansion framework that aims to 
dynamically expand a seeded taxonomy horizontally 
(width expansion) and vertically (depth expansion) and performs entity linking 
with Microsoft's Probase \cite{wu2012probase} ---  a probabilistic framework used to harness 2.7 million concepts 
mined from 1.68 billion web pages --- to iteratively grow a seeded taxonomy. 
As an entity is matched against a verified knowledge base, we perceive the accuracy of terms and concept relations  to be higher than that of CoRel and TaxoGen. 
 
Authors of the HiExpan, as well as some volunteers assessed the taxonomy parent-child pair relations
using ancestor- and edge-F1 scores.


\subsection{Observations}

Each of those taxonomy extractors face their own set of advantages and drawbacks. 
TaxoGen  is the only parameterized systems in our experiments, and is the only one that does not require 
a seeded input for producing an output, 
which can be beneficial when prior knowledge of the corpus is lacking. It also  generates alternative synonyms for each taxonomy topic, which increases the coverage and improves concept mapping between taxonomies and documents.
However, it seems to depend on the keyword extraction quality and
it is unclear how to determine the best hyper-parameter settings
owing to the lack of automatic evaluation methods. 

CoRel uses the concept pairs provided in the seed taxonomy  for mining similar relations, but this has become 
its Achilles' heel because same-sentence co-occurrence of valid parent-child topics
is rare in real-world data. As a result, CoRel may fail to produce any output at all due to insufficient
training examples for the relation classifier. It is also 
resource-intensive for making use of neural networks for relation transferring and depth expansion.
Anecdotally, the output of CoRel may also not be entirely exhaustive and deterministic.
 
For our experiments, HiExpan is perceived to produce the most consistent taxonomies
thanks to the use of Probase for measuring topic similarities and locating related concepts. 
However, the set-expansion mechanism of HiExpan often ignores topic granularity and 
adds hyponyms and hypernyms found in similar contexts to the exact same taxonomy level
(hence most HiExpan taxonomies are two-level only). It also cannot differentiate word senses such as virus as in \emph{computer virus} and a \emph{viral disease}.

\begin{table*}[t!]
 \begin{center}
    \begin{tabular}{@{}llllllr@{}}
    \toprule
    \textit{c}                                & Pred 1    & Pred 2 & Pred 3    & Pred 4  & Pred 5  & Rank \\ \midrule
    Mussel  & fish (0.227)    & dish (0.144) & \hl{seafood (0.140)} & meat (0.037)  & soup (0.033)  & 3    \\
    Clam     & fish (0.203)    & dish (0.095) & \hl{seafood (0.076)} & crab (0.030)  & thing (0.027) & 3    \\
    Lobster  & \hl{seafood (0.222)} & dish (0.145) & lobster (0.131) & food (0.052)  & sauce (0.052)  & 1    \\
    Chicken & dish (0.167)    & meat (0.110) & chicken (0.079) & thing (0.058) & sauce (0.052) & 73   \\
    Beef       & meat (0.274)    & beef (0.161) & dish (0.063)    & food (0.027)  & thing (0.024) & 57   \\ \bottomrule
    \end{tabular}
    \caption{Top-5 hypernym predictions made by  a pre-trained BERT model (Bert-large-uncased-whole-word-masking) by prompting it with ``\textit{c is a type of [MASK]}''. The rank of seafood in the list is indicated in the last column.}
    \label{tab:hypernymy_recall}
 \end{center}
  \end{table*}

\section{\rate}
\label{sec:approach}

A critical part of taxonomy/ontology evaluation is knowledge about subsumptions, e.g.
"is \emph{fluorescence spectroscopy} a type of \emph{fluorescence technology}?" or "is \emph{CRJ200} 
a \emph{Bombardier}?".

Thus, \rate{} measures the  accuracy of the hypernym relations present in a taxonomy we seek  to evaluate. The main difference between our work and earlier ones is that we do not rely on human judgments to determine the quality of a parent-child pair, nor do we consider an external reference (that often is not available or simply too shallow). Instead, we rely on a large language model  tasked to check subsumption relations. 

Ultimately, an optimized language model should be able to generate an accurate list of 
the most canonical hypernyms for a given domain, similar to domain experts.
But because we are mainly interested in domain-specific taxonomies, there is a high risk that specific terms of the domain are not well recognized by the model, and therefore, we investigate  three methods for increasing the hit rate of hypernymy prediction of taxonomy subjects and reducing false negatives by (1) creating various prompts, 
(2) fine-tuning MLMs with different masking procedures, and (3) extending the model's vocabulary with concept names.

\subsection{Core idea}

We consider a  taxonomy as a set of $n$ parent-child pairs from adjacent taxonomy levels linked by single edges, denoted as $(p,c) \in \mathcal{T}$. For each parent-child pair $(p_i,c_i), i \in 1,...,n$, 
we insert $c_i$ and the "[MASK]" token into prompts containing 
"is-a" patterns  \cite{hearst1992automatic}, then use LMs to unmask $p'_1(c_i), p_2'(c_i), ..., p_k'(c_i) \in p'(c_i)$ per query 
as proxy parent terms of $c_i$, where $k$ is a recall threshold (we used $k=10$ in this work). This process is illustrated in Table~\ref{tab:hypernymy_recall}. 

A good pair of taxonomy concepts is therefore if the parent concept $p_i$ can be found among the machine predictions $p’(c_i)$. We consider a parent-child relation \emph{positive} if and only if the parent term 
is recalled one or more\footnote{A parent word can be predicted multiple times in singular and plural forms,  misspellings, and so on,
e.g. "dessert", "desserts" and "desert".} times in the top $k$ predictions. This policy can obviously be adjusted, which we leave as future work.
The measure of quality of $\mathcal{T}$ is then simply the percentage of $(p,c)$ links in $\mathcal{T}$ that are correct according this procedure. We note that for a taxonomy with no parent-child pairs, i.e. a single-level taxonomy, our evaluation score is $0$.

 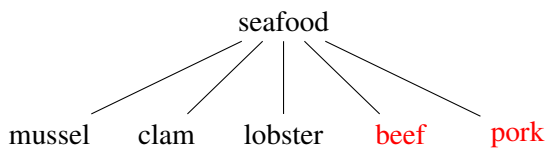
\begin{figure}[htp!]
 \begin{center}
  \begin{tikzpicture}
  [
      sibling distance=4em
  ]
      \node {seafood}
          child {node {mussel}}
          child {node {clam}}
          child {node {lobster}}
          child {node {\textcolor{red}{beef}}}
          child {node {\textcolor{red}{pork}}};
  \end{tikzpicture}
  \caption{Excerpt from HiExpan1 for topic "seafood"}
  \label{fig:hiexpan1_seafood}
 \end{center}
\end{figure}

As an illustration, the taxonomy in Figure~\ref{fig:hiexpan1_seafood} would receive a score of 3/5 based on the predictions made in Table \ref*{tab:hypernymy_recall} where for instance,  $p'_1(c_i), p_2'(c_i), ..., p_5'(c_i)$ equal \emph{fish, dish, seafood, meat, soup} for $c_i = \textit{mussel}$, in which we find the real taxonomy parent $p_i = \textit{seafood} = p_3'(c_i)$.



We observe from  Table \ref*{tab:hypernymy_recall} that not every prediction
is factually correct (e.g. mussels are neither fish nor meat), and it remains evidently unreliable 
to depend solely upon pre-trained language models as ground-truth for all knowledge domains. Yet,  we argue that we can regard the rankings of MLM predictions
as a likelihood of a subsumption relation between the subject and the object of a query. In our example, the model is significantly more likely to predict
``seafood'' for \emph{mussel, clam} and \emph{lobster} (rank 3,3,1) than for \emph{chicken} and \emph{beef} (rank 73,57).

\begin{table*}[tp!]
\resizebox{\textwidth}{!}{%
\begin{tabular}{lllllllr}
\hline
\multicolumn{2}{l}{Prompt}                     & Pred1      & Pred2    & Pred3      & Pred4       & Pred5          & Rank \\ \hline
p1a  & \{shrimp\} {[}MASK{]}                  & salad      & cocktail & pasta      & soup        & rice           & 359  \\
p1b  & {[}MASK{]} \{shrimp\}                  & fried      & no       & garlic     & coconut     & fresh          & 117  \\ \hline
p2a  & \{shrimp\} is a {[}MASK{]}             & joke       & must     & winner     & favorite    & hit            & 959  \\
p2b  & \{shrimp\} is an {[}MASK{]}            & option     & issue    & experience & art         & order          & 4407 \\ \hline
p3a  & \{shrimp\} is a kind of {[}MASK{]}     & joke       & thing    & dish       & treat       & disappointment & 146  \\
p3b  & \{shrimp\} is a type of {[}MASK{]}     & dish       & thing    & food       & sauce       & \hl{seafood}        & 5    \\
p3c  & \{shrimp\} is an example of {[}MASK{]} & that       & this     & shrimp     & food        & \hl{seafood}        & 5    \\ \hline
p4a  & {[}MASK{]} such as \{shrimp\}        & sides        & food  & \hl{seafood}       & fish & shrimp        & 3   \\
p4b  & A {[}MASK{]} such as \{shrimp\}        & lot        & variety  & side       & combination & protein        & 40   \\
p4c & An {[}MASK{]} such as \{shrimp\}       & ingredient & item     & option     & order       & animal         & 197  \\ \hline
p5a & My favorite {[}MASK{]} is \{shrimp\}   & dish       & thing    & part       & item        & roll           & 16   \\ \hline
\end{tabular}%
}
\caption{Evaluation queries for the parent-child pair (seafood,shrimp).}
\label{tab:seafood-shrimp}
\end{table*}

\subsection{Diversified Prompting}

Models can produce all sorts of trivial predictions, such as stop-words (e.g. "\textbf{this} is a kind of seafood"), or
expressions and collocations found frequently in training samples 
(e.g. "seafood is a kind of \textbf{joke/disappointment}").

Differences in prompts used can actively impact a model's performance in hypernymy retrieval 
\cite{peng2022discovering,hanna2021analyzing}.  \citet{hanna2021analyzing} reported that prompting BERT for hypernyms can actually outperform other unsupervised methods even in an unconstrained scenario, but the  effectiveness of it depends on the actual queries. For example, they show that the query ``A(n) $x$ \textbf{is a} [MASK]'' outperformed ``A(n) $x$ \textbf{is a type of} [MASK]'' on the Battig dataset.

As a result, instead of relying on a single query, we design five pattern groups (p1-p5) of hypernymy tests
for pooling unmasking results. Those are illustrated in Table \ref*{tab:seafood-shrimp}
for the  parent-child pair (seafood,shrimp).   

While p2 to p4 follow standard Hearst-like patterns \cite{hearst1992automatic}, 
p5a employs the ``my favourite is'' prompt which has demonstrated high P@1 and MRR in \cite{hanna2021analyzing}.
Patterns p1 have been created specifically for noun phrases that have a tendency to be split and
considered as good taxonomy edges by ATC systems.\footnote{For instance, extractors tend to produce  (salad,shrimp) for the pair (salad,shrimp salad).}
 
With this refined set of patterns, a topic pair has therefore a score of 1, as in the seafood-shrimp example,
if the parent term is among the top-k machine predictions for any inquiries containing the child topic, 
and 0 vice versa. Again, more elaborate decisions can be implemented. 

\subsection{Fine-tuning the Language Model}

To improve hypernymy predictions, we must also address two issues with pre-trained language models:
(1) the models are untrained on the evaluation domain; (2)
the default model tokenizer and vocabulary are oblivious of some taxonomy topics, resulting in lower recall.

Most research on MLM prompting only assessed the
performance of pre-trained models.  Yet, \citet{peng2022discovering} found an improvement 
when using FinBert models \cite{yang2020finbert} pre-trained with massive financial corpora 
in retrieving financial hypernyms such as \emph{equity} and \emph{credit} for 
\emph{``S\&P 100 index is a/an \_\_ index"}, compared to using BERT-base.
Also, \citet{dai2021ultra} generated ultra-fine entity typing labels, e.g. 
``person, soldier, man, criminal" for \emph{``\textbf{he} was confined at Dunkirk, escaped, set sail for India"}
through inserting hypernym extraction patterns and training LMs to predict such patterns.

Analogously, we compared six fine-tuned models, investigating different masking protocols, model vocabulary (see next section) and training sizes.
Because we want the language models to concentrate on the taxonomy entities, particularly the parent terms 
and their surrounding contexts, we prioritize therefore masking the main topics (shown  in Table~\ref*{tab:mlm-main-topics}) and parent terms of the taxonomies to evaluate, then other taxonomy entities (e.g. leaf nodes), 
followed by AutoPhrase entities if no taxonomy entities are present in the sentence 
and other random tokens from our training samples.
In addition, we test entity masking by only masking \emph{one} taxonomy entity rather than 15\% of sentence tokens
to gain more sentence contexts. Our masking procedures are illustrated in Figure~\ref*{fig:masking-logic}.

\subsection{Extended Vocabulary}
\label{sec:extended_voc}

Domain-specific words such as food items are typically not predicted as a whole word, but rather as a sequence of 
subword units, such as \emph{appetizer} which is treated as \emph{'app', '\#\#eti' and '\#\#zer'} by the standard
tokenizer. To avoid multi-unit words to be overlooked by the language model, we propose to extend its vocabulary.  

\begin{figure}[ht]

\begin{tabular}{@{}l|p{6.5cm}|@{}} 
\cline{2-2}
\multirow{3}{*}{\rotatebox[origin=c]{90}{Review}}   &                                                           
Everything was pretty good but the \entity{beef} in the \entity{mongolian beef} was very \entity{chewy} and had a \entity{weird texture}.\\              \cline{2-2}
\end{tabular}
 
\begin{tabular}{@{}l|lp{4.3cm}|@{}}
\cline{2-3}  
\multirow{4}{*}{\rotatebox[origin=c]{90}{Entities}}   &                                                           
  Taxonomy  & beef (CoRel1-4, HiExpan1)\\
  &                  & mongolian (CoRel1-4)    \\ 
  & AutoPhrase  & beef, chewy, mongolian \\
                     & & weird texture                                                                      \\ \cline{2-3}
\end{tabular}

\vspace*{2mm}

\hspace*{3cm}Masking Policy

\begin{tabular}{@{}l|lp{5.32cm}|@{}}
\cline{2-3}  
\multirow{8}{*}{\rotatebox[origin=c]{90}{Entity}}   
&  15\%  & Everything was pretty good but the {[}MASK{]} in the {[}MASK{]} {[}MASK{]} was very chewy and had a weird texture.\\
&  one &  Everything was pretty good but the {[}MASK{]} in the mongolian {[}MASK{]} was very chewy and had a weird texture.\\ \cline{2-3}
\end{tabular}

\begin{tabular}{@{}l|lp{5.32cm}|@{}}
\cline{2-3}   
\multirow{4}{*}{\rotatebox[origin=c]{90}{Token\textcolor{white}{y}}}
& 15\% & Everything was pretty {[}MASK{]} but the beef in the mongolian beef {[}MASK{]} very chewy and had a {[}MASK{]} texture.\\
\cline{2-3}
\end{tabular}
 
\caption{Comparison of masking strategies for a sample Yelp review where taxonomy entities or those proposed by AutoPhrase are underlined. We prioritize masking the taxonomy entities,
AutoPhrase entities and random tokens, in that order. }
\label{fig:masking-logic}

\end{figure}


We enrich the vocabulary of models m1 and m2, by adding the lemmas (or singular forms) of 
parent terms from Table~\ref*{tab:mlm-main-topics} that were not previously included in the
base tokenizer, such as ``sushi", ``appetizer" and ``carne asada",
and resizing the models' token embedding matrices to match the size of the new tokenizer.
The embedding representation of new tokens were initialized randomly before fine-tuning, although it is possible to 
assign them to the representation of the closest terms in the original vocabulary.

 By adding only a small number of new tokens to the model and tokenizer,
we also ensure similar model and tokenizer efficiencies. We believe that vocabulary extension will become a necessary 
step for effective hypernymy prediction in most specialized domains, 
though the exact optimal strategies remain to be discussed.

\section{Experiments}
\label{sec:experiments}
 
We conducted our experiments on the Yelp corpus which contains around 1.08M restaurant reviews such as the one in Figure~\ref{fig:masking-logic} (top box).  We used the very same corpus prepared by \citet{huang2020corel}.\footnote{Available at: \url{https://drive.google.com/drive/folders/13DQ0II9QFLDhDbbRcbQ-Ty9hcJETbHt9}.}

\subsection{Taxonomies} 

With the ATC systems described in Section~\ref{sec:system_description} we produced seven taxonomies we seek to evaluate.  Our objective here was to explore the extractors so to get the best taxonomies in a fair amount of explorations. 
For TaxoGen, we only had to  specify some parameters.\footnote{We considered taxonomy depth, number of topics per level, and “word filtering threshold”. See the github for the specific values we used.} For CoRel and HiExpan however, we had to provide a seed taxonomy. Hence we produced 5 such taxonomies,\footnote{They pretty much align with the one used by  \citet{huang2020corel}, although we proceeded by trial-error  until satisfaction.} mainly  seeking for frequent parent-child pairs.   
 
Table~\ref{tab:mlm-main-topics} reports the main topics (level 1) of the resulting taxonomies. We observe that the output of one ATC system varies substantially from one parametrization to another. Also, it is noticeable that main topics lack some structure. For instance, grouping \emph{beef} and \emph{pork} into the a category \emph{meat} would arguably make sense in the output of CoRel1.

\begin{table}[t]
  \begin{center}

    \begin{tabular}{@{}p{7.5cm}@{}}
    \toprule
    \makebox[2.2cm][l]{Taxonomy}   Top level (main) topics                                                     \\ \midrule
    \fespace{CoRel1}      steak, veggies, beef, cheese, crispy, fish, rice, salad, shrimp, spicy, pork, bacon, burger, appetizer, bread, dessert, seafood\\ 
     \fespace{CoRel2}      bacon, bread, fries, roll, soup, burger, dessert, salad, shrimp \\
    \fespace{CoRel3}      chinese, seafood, dessert, steak                                \\
    \fespace{CoRel4}      dinner, food, location, lunch, service                          \\
    \fespace{HiExpan1}  seafood, salad, dessert, appetizer, food, sushi, desert, pizza, coffee, bread, pasta, beer, soup, wine, cheese, cocktail, taco, water, music \\
    \fespace{TaxoGen1}   main\_dish, south\_hills, high\_cei\-lings, était\_pas           \\
    \fespace{TaxoGen2}  chest, tempe, amaretto, pepper\_jelly, relies, travis, free\_admission, exposed\_brick\\ \bottomrule
    \end{tabular}%
 \end{center}
    \caption{Main targets of MLM evaluation.}
    \label{tab:mlm-main-topics}
\end{table}

\subsection{Models}

We fined-tuned six  language models according to the different strategies we presented in Section~\ref{sec:approach} and which characteristics  are summarized in Table~\ref{tab:mlm-models}.  In particular, we experiment with \emph{entity masking} while fine-tuning model m1a, m1b and m0b, which emphasizes masking task-relevant tokens, because it has been shown to be more effective than \emph{random masking} in \cite{sun2019ernie,kawintiranon2021knowledge}.  All models have been fine-tuned for 2 epochs by masking 15\% tokens, to the exception of m1b (marked with $\star$) for which only one entity has been masked per example.

\begin{table}[ht]
\setlength{\tabcolsep}{3pt}
\begin{tabular}{@{}llllll@{}}
\toprule
\multicolumn{1}{l}{Model}                                              & \multicolumn{3}{l}{Finetuning}           & \multicolumn{2}{c}{Masking}\\ 
name (base)                                                                   & Voc.               & Full & 70\%                                        & Ent. & Tok                                                                     \\ \midrule
m1a    (\href{https://huggingface.co/bert-base-uncased}{bert-base})                                                                &  \checkmark   & \checkmark &                                    & \checkmark             &  \\
m1b     (\href{https://huggingface.co/bert-base-uncased}{bert-base})                                                                & \checkmark    &   \checkmark & 				  & \checkmark$\star$  & \\
m2a    (\href{https://huggingface.co/bert-base-uncased}{bert-base})                                                                 &  \checkmark   &  \checkmark &  				  &    			            & \checkmark    \\
m2b    (\href{https://huggingface.co/bert-base-uncased}{bert-base})                                                                & \checkmark     &                     & \checkmark  		  & 			           & \checkmark     \\ \midrule
m0a     (\href{https://huggingface.co/bert-base-uncased}{bert-base})                                                              &                          &  		      & \checkmark                & 				   & \checkmark     \\
m0b   (\href{https://huggingface.co/distilbert-base-uncased}{distilbert-base})   &          &   & \checkmark   		 & \checkmark             &  \\ \midrule
\end{tabular}%
\caption[Evaluation model configurations]{Configurations of the fine-tuned models,
with models m0a and m0b serving as baselines for training with the base tokenizer; m0b using a smaller pre-trained model and less fine-tuning material. Column Voc. indicates that main target words proposed  ATC systems were injected in the model's vocabulary.}
\label{tab:mlm-models}
\end{table}

For comparison purposes, we also selected two pre-trained models  \href{https://huggingface.co/bert-large-uncased-whole-word-masking}{bert-large-uncased-whole-word-masking} and \href{https://huggingface.co/bert-base-uncased}{bert-base-uncased} 
that we did not fine-tune and that we named  B-l and B-b respectively.

To highlight the qualitative differences between our evaluation models, 
we provide a simple prompt ``my favorite [MASK] is sirloin" for the models to predict the taxonomy hypernym "steak" in CoRel1. The results are shown in Table~\ref{tab:sirloin}, where 5 out of 6 fine-tuned models and none of the pre-trained models correctly predicted the taxonomy parent in the top 4 predictions. 
Further, all fine-tuned models returned "steak" in the top ten predictions. 

\begin{table}[t]
\resizebox{\columnwidth}{!}{%
\begin{tabular}{@{}lllllr@{}}
\toprule
 Model     & Pred1 & Pred2  & Pred3  & Pred4    & Rank \\ \midrule
  m1a &
  burger &
  dish &
  sandwich &
  \hl{steak} & 
    4 \\
 m1b   & dish  & burger & beer   & sandwich     & 10   \\
  m2a   & \hl{steak} & dish   & meat   & cut         & 1    \\
m2b   & \hl{steak} & dish   & burger & meat          & 1    \\
m0a    & dish  & burger & \hl{steak}  & meat     & 3    \\
m0b   & cut   & \hl{steak}  & meat   & beef        & 2    \\ \midrule
B-l &
  fruit &
  flavor &
  food &
  color & 
  69 \\
B-b & food  & drink  & color  & dessert    & 71   \\ \bottomrule
\end{tabular}%
}
\caption{Fine-tuned (top) vs. pre-trained (bottom) models' top-4 predictions with the prompt ``my favorite {[}MASK{]} is sirloin ."}
\label{tab:sirloin}
\end{table}

Lastly, we show the positive effects of extending the vocabulary of the language model  in Table~\ref{tab:mozzarella}
where we wish to recall the parent term ``appetizer" for the concept pair ``appetizer-mozzarella sticks" 
in CoRel1, where the token ``appetizer" would be split into \emph{'app', '\#\#eti' and '\#\#zer'} by the standard
tokenizer. Both models m1a and m1b trained with entity masking and an expanded vocabulary  
correctly predicted ``appetizer" in their top five predictions; m2 models also recalled the term, albeit with
a very low rank whereas other models are completely oblivious to it.
Nevertheless, we find that expanding the model's vocabulary
in conjunction with entity masking may introduce bias into the models when fine-tuning with limited training samples, 
i.e. over predicting the added tokens.

\begin{table}[t]
 \resizebox{\columnwidth}{!}{%
\begin{tabular}{@{}lllllr@{}}
\toprule
Model                                                      & Pred1  & Pred2  & Pred3     & Pred4        & Rank \\ \midrule
m1a                                                     & sides  & foods  & food      & apps       & 5    \\
m1b                                                    & sides  & food   & \hl{appetizer} & foods    & 3    \\
  m2a                                                     & sides  & items  & food      & dessert     & 6089 \\
 m2b                                                     & things & items  & foods     & props      & 3111 \\ \midrule
m0a                                                     & sides  & extras & items     & dessert    & N/A  \\
m0e                                                    & sides  & apps   & foods     & snacks     & N/A  \\ 
 B-l                                                  & foods  & items  & products  & food       & N/A  \\
 B-b                                                  & foods  & snacks & food      & items    & N/A  \\ \bottomrule
\end{tabular}%
 }
\caption{Top-4 predictions of models with extended  (top) or base  (bottom) vocabulary for the prompt ``{[}MASK{]} such as mozzarella sticks".}
\label{tab:mozzarella}
\end{table}

\begin{table*}[htp!]
\begin{center}
\begin{tabular}{@{}lrrrrrr|rr|rc|c@{}}
\toprule
           & \multicolumn{6}{c}{Fine-tuned Models} & \multicolumn{2}{|c|}{BERT} & Majority & \rate & Manual\\ 
               & m1a & m1b & m2a & m2b & m0a & m0b & large & base & Voting & ranking & ranking \\ \midrule
CoRel1   & 72.7   & 71.8   & 42.4   & 44.5   & 46.3   & 43.6   & 20.4      & 27.4     & 44.3  			& 4 & 3  \\
CoRel2   & 78.2   & 75.0   & 54.4   & 53.7   & \textbf{57.2}   & 51.2   & 25.9      & 36.2     & 57.2        & 2 & 2 \\
CoRel3   & 60.2   & 66.7   & 54.1   & 54.9   & \textbf{57.2}   & 50.1   & 36.0      & 40.0     & 53.5    	& 3 & 4 \\
CoRel4   & 68.2   & 64.6   & 45.0   & 39.0   & 36.5   & 38.1   & \textbf{41.0}      & 41.8     & 34.7        & 5 & 5 \\
HiExpan1 & \textbf{84.5}   & \textbf{84.7}   & \textbf{59.5}   & \textbf{56.7}   & 56.9   & \textbf{64.3}   & 34.9      & \textbf{42.0}     & \textbf{59.0}    & 1 & 1 \\
TaxoGen1 & 13.5   & 14.7   & 5.5   & 6.1   & 1.2   & 2.5   & 3.1      & 3.7     & 1.2    & 6 & 6 \\
TaxoGen2 & 0.0   & 0.0   & 0.0   & 5.0   & 0.0   & 0.0   & 0.0      & 0.0     & 0.0    & 7 & 7 \\ \bottomrule
\end{tabular}%
\end{center}

\caption[Automatic relation accuracy scoring]{Relation accuracy scores evaluated by language models,
calculated by the number of positive relations, or parent terms in the model predictions, divided by the number of 
unique parent-child pairs in each taxonomy.}
\label{tab:ra-score}
\end{table*}

\subsection{Ranking Results}
 
\subsubsection{Manual Ranking} 

The first author of this paper manually ranked the taxonomies we built, prior to experimenting \rate. The main task has been to manually verifying the quality of the parent-child pairs of each taxonomy, while also taking into account factors like taxonomy structure.\footnote{All parent-child pairs of HiExpan1 and TaxoGen1\&2 have been evaluated, but we sampled concept pairs from CoRel 1-4 because their word clusters are too large.}

HiExpan1 was deemed the best likely because word relations actually come from a verified database and we found the coverage to be  broad. It is also observably more accurate than CoRel 1-4, which have similar (overall good) quality. 
TaxoGen taxonomies were the least accurate (TaxoGen1 being better than TaxoGen2). 
We found them trivial, in that many unimportant topics are picked by the algorithm. One reason for this we believe is the sensitivity of the system to the keywords generated by AutoPhrase, which on Yelp is generating too many irrelevant terms, leading to many noisy pairs (e.g. “exposed brick – music video”). 

In fine, it was easy and straightforward to rank the HiExpan and TaxoGen taxonomies but more difficult for rank the CoRel taxonomies. Such an evaluation is delicate, after all, this was the main motivation for this study.

 \subsubsection{\rate{} Ranking} 

Table \ref*{tab:ra-score} showcases the results of MLM taxonomy relation accuracy evaluation, 
calculated by the number of positive relations over all unique parent-child pairs in a taxonomy.\footnote{We considered word inflections and certain special cases to improve matching between taxonomy terms and 
machine predictions, e.g. "veggies", "vegetable" and "vegetables"; "dessert" and "desert".}

The entity-masking models m1a and m1b predicted the most positive relationships in each candidate
taxonomy while the pre-trained models predicted the fewest, which was expected.  It is also surprising that B-b outperforms B-l when it comes to matching more positive concept pairs. Model m2b (trained on two-thirds of the data) expectedly underperforms model m2a, but not drastically. 

However, all models produce overall similar score distributions, with the HiExpan taxonomy receiving
the highest scores and the TaxoGen taxonomies receiving the lowest. This is consistent with our 
manual judgements in that the HiExpan concept pairs were derived from an accurate relation dataset
(Probase), whereas TaxoGen1 and TaxoGen2 contain mostly noise.

We also compute the majority voting scores for each evaluation target
using the six  models of Table~\ref{tab:mlm-models}: 
a concept pair of a taxonomy is positive
if and only if three or more models have successfully predicted the parent word. The resulting ranking is reported in the next column, and is shown to correlate well with our manual evaluation (last column).

%
  
\subsection{Random noise Simulation}

\begin{figure}[h]
 
 \begin{center}
\includegraphics[width=0.85\columnwidth]{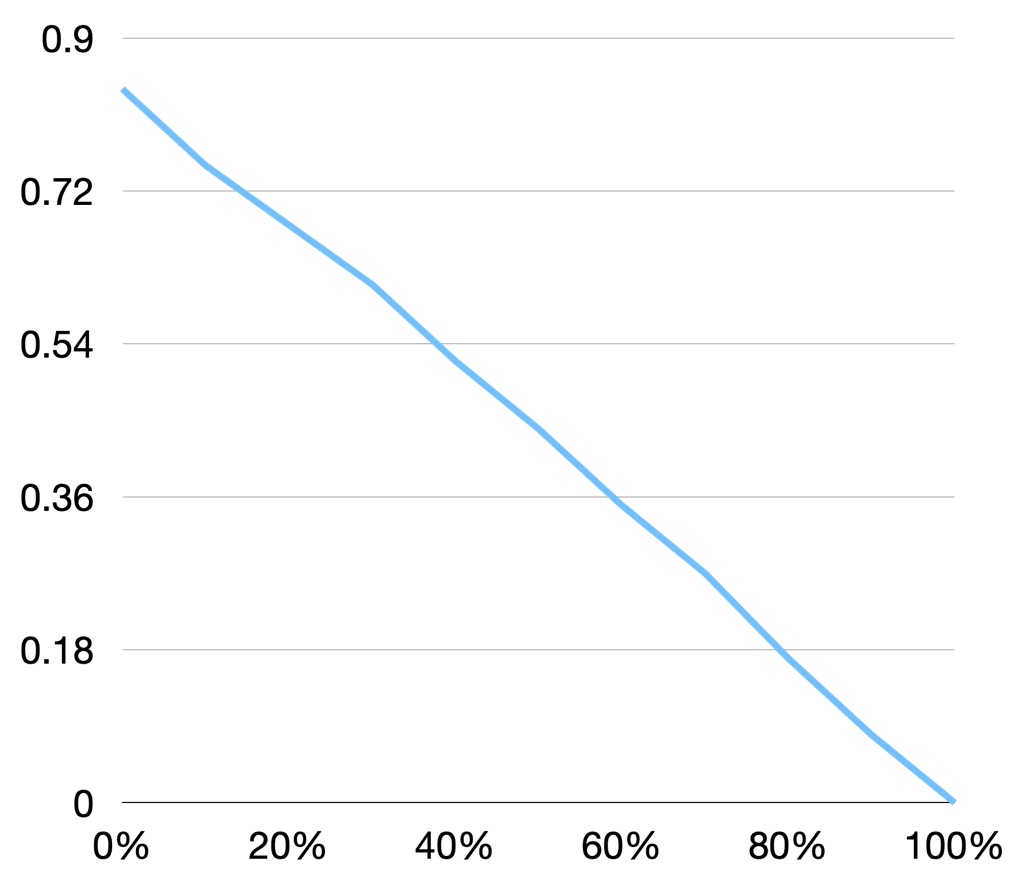}
 \end{center}
\caption{Relation accuracy obtained with model m1a, as a function of the percentage of noise introduced in HiExpan1.}
\label{fig:noise}

\end{figure}

To further evaluate the good behaviour of \rate, we conducted an experiment  where we degraded the HiExpan1 taxonomy (the best one we tested). We did this by randomly replacing a percentage of concepts by others. Figure~\ref{fig:noise} shows that the score (obtained with model m1a) roughly decreases linearly with the level of noise introduced, which is reassuring. 

%
%

\section{Discussion}
\label{sec:discussion}

 We presented \rate, a procedure  aimed at automatically evaluating a domain taxonomy without reference taxonomies or human evaluations. It relies on a large language model and an unmasking procedure for producing annotations.   
We tested \rate{} on the Yelp corpus which gathers restaurant reviews, and found that it well behaves: it correlates well with human judgments, and (artificially) degrading a taxonomy leads to a score degradation proportional to the amount  of noise injected. Still, we observed that the quality of the language model predictions varies according to the strategies used to fine-tune them.    
 
There remains a number of avenues to investigate. First, we have already identified a number of decisions that could be revisited. In particular, we must test \rate{} on other domains, possibly controlling variables such as the size of the fine-tuning material or the frequency of terms.  Second, \rate{} is an accuracy measure, and depending on the evaluation scenario, it  should eventually be coupled with a measure of recall.  Last, an interesting avenue is to investigate whether  \rate{} can be used to optimize the hyper-parameters of an ATC system. 
 
\section*{Acknowledgments}

This work has been done in collaboration with IATA to whom we are grateful. 

\bibliographystyle{acl_natbib}
\bibliography{ref}

%
%

\end{document}